\documentclass[sigconf]{acmart}
\AtBeginDocument{%
  \providecommand\BibTeX{{%
    \normalfont B\kern-0.5em{\scshape i\kern-0.25em b}\kern-0.8em\TeX}}}

\usepackage{mathtools} 
\usepackage{booktabs} 
\usepackage{tikz} 
\usepackage{algorithm}
\usepackage{algorithmic}

\newtheorem{definition}{Definition}
\usepackage{subfigure}
\usepackage{xcolor}
\usepackage{amsmath}
\usepackage{amsfonts}
\usepackage{todonotes}

\copyrightyear{2022} 
\acmYear{2022} 
\setcopyright{acmcopyright}\acmConference[CIKM '22]{Proceedings of the 31st ACM International Conference on Information and Knowledge Management}{October 17--21, 2022}{Atlanta, GA, USA}
\acmBooktitle{Proceedings of the 31st ACM International Conference on Information and Knowledge Management (CIKM '22), October 17--21, 2022, Atlanta, GA, USA}
\acmPrice{15.00}
\acmDOI{10.1145/3511808.3557672}
\acmISBN{978-1-4503-9236-5/22/10}
\begin{document}

\title{Probabilistic Model Incorporating Auxiliary Covariates to Control FDR}

\author{Lin Qiu}
\email{lin.qiu.stats@gmail.com}
\affiliation{%
  \institution{The Pennsylvania State University}
  \city{State College}
  \state{PA}
  \country{USA}
}

\author{Nils Murrugarra-Llerena}
\email{nmurrugarrallerena@weber.edu}
\affiliation{%
  \institution{Weber State University}
  \city{Ogden}
  \state{UT}
  \country{USA}
}

\author{Vítor Silva}
\email{vitor.silva.sousa@gmail.com}
\affiliation{%
  \institution{Snap Inc.}
  \city{Santa Monica}
  \state{CA}
  \country{USA}
}

\author{Lin Lin}
\email{l.lin@duke.edu}
\affiliation{%
  \institution{Duke University}
  \city{Durham}
  \state{NC}
  \country{USA}
}

\author{Vernon M. Chinchilli}
\email{vchinchi@psu.edu}
\affiliation{%
  \institution{The Pennsylvania State University}
  \city{Hershey}
  \state{PA}
  \country{USA}
}

\renewcommand{\shortauthors}{Lin Qiu et al.}

\begin{abstract}
Controlling False Discovery Rate (FDR) while leveraging the side information of multiple hypothesis testing is an emerging research topic in modern data science. Existing methods rely on the \textit{test-level covariates} while ignoring
metrics about \textit{test-level covariates}.
This strategy may not be optimal for complex large-scale problems, where indirect relations often exist among \textit{test-level covariates} and \textit{auxiliary} metrics or covariates. We incorporate \textit{auxiliary covariates} among \textit{test-level covariates} in a deep Black-Box framework 
(\texttt{named as NeurT-FDR}) which boosts statistical power and controls FDR for multiple hypothesis testing. Our method parametrizes the \textit{test-level covariates} as a neural network and adjusts the \textit{auxiliary covariates} through a regression framework, which enables flexible handling of high-dimensional features as well as efficient end-to-end optimization. We show that \texttt{NeurT-FDR} makes substantially more discoveries in three real datasets compared to competitive baselines.
\end{abstract}

\ccsdesc[500]{Mathematics of computing~Probabilistic algorithms}

\keywords{Social Media Content Understanding, Multiple Hypothesis Testing, FDR Control}


\maketitle

\section{Introduction}
In modern statistics, from genetics, neuroimaging, to online advertising, researchers routinely test thousands or millions of hypotheses at a time~\cite{zhang19} to discover unique data instances. Current approaches \cite{efron04} solve this problem via Multiple Hypothesis Testing (MHT). MHT aims to maximize the number of discoveries while controlling the False Discovery Rate (FDR). For example, in social media, we may want to identify popular social media posts than normal ones. Also, in biology, we may want to discover which cancer cells respond positively to the treatment under a new drug.

\begin{figure}[ht] 
    \centering
    \includegraphics[width=0.45\textwidth]{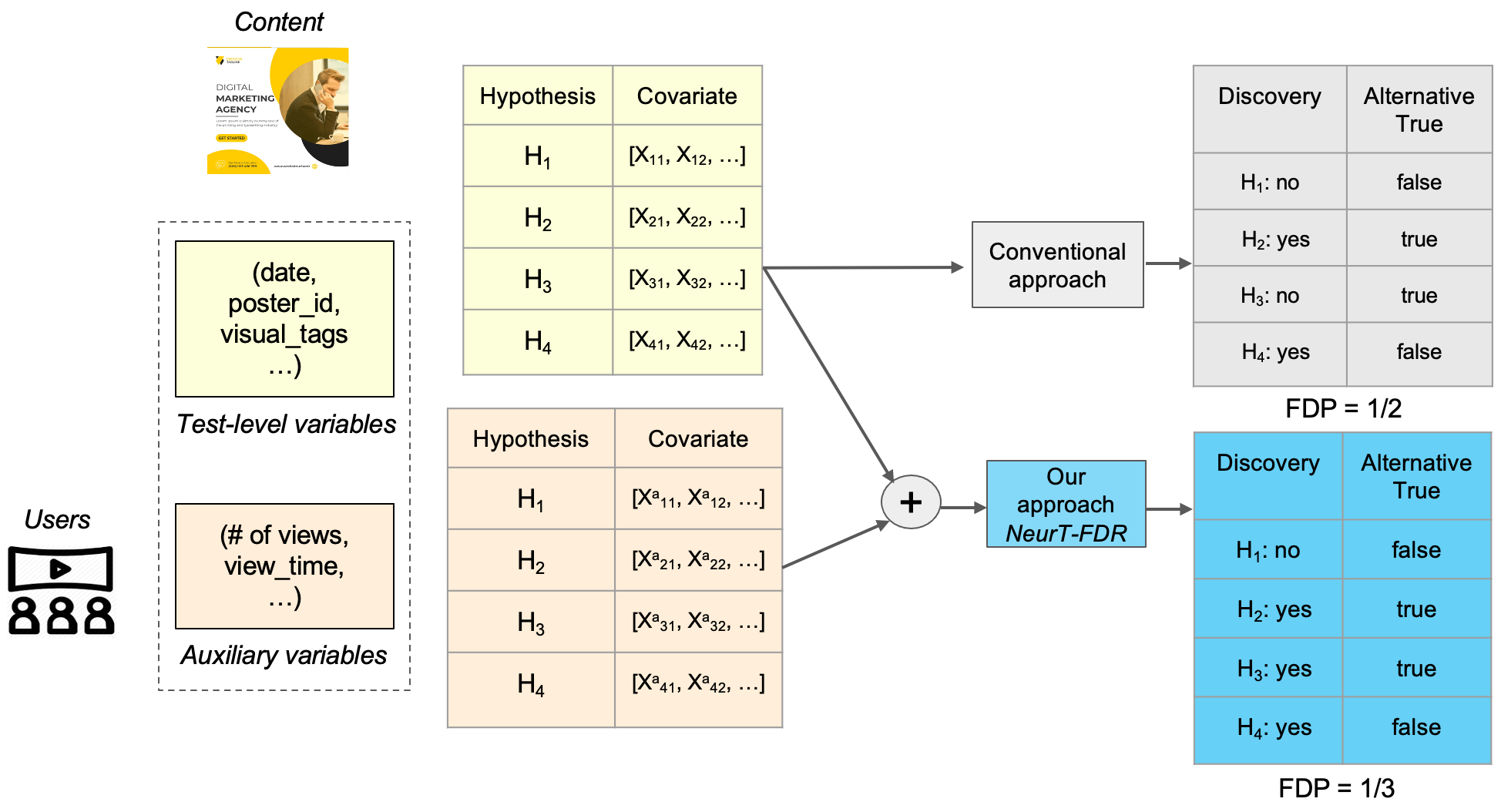}
    \caption{Each hypothesis has \textit{test-level covariates} and \textit{auxiliary covariates}. Existing covariate-adaptive FDR methods consider only \textit{test-level covariates}, while we propose a new method (\textit{NeurT-FDR}) to incorporate both \textit{test-level covariates} and \textit{auxiliary covariates} via a neural network.
    }
    \label{fig:concept}
\end{figure}

Existing MHT approaches \cite{tansey18, xia17, zhang19} only use covariate-adaptive FDR procedures on top of \textit{test-level covariates} to improve the detection power while maintaining the target FDR. \textit{Test-level covariates} only provide characteristics of the samples in the dataset, which can be metadata of social media posts, or genomic profiles for each cell. However, depending on the domain, we can access complementary information besides \textit{test-level covariates} that can facilitate the work of MHT approaches. For example, as shown in Figure \ref{fig:concept}, in the social media domain, the goal is to find engaging content, and the post can be represented by visual tags and metadata information. Additionally, content consumption metrics, such as the number of views and content view time, are available. These metrics encapsulate information that facilitates MHT work. This additional information is called \textit{auxiliary covariates} and corresponds to the samples in the dataset. More specifically, content consumption metrics do not correspond to characteristics of the sample, i.e., posted content, but how users interact in the platform to access this content. Typically, such \textit{auxiliary covariates} are of lower dimension than those \textit{test-level covariates} (e.g., visual tags), and are more structured.

In this paper, we present a hierarchical probabilistic black-box method which incorporates  \textit{test} and \textit{auxiliary covariates} to control the FDR, named \textit{NeurT-FDR}. Our main contributions can be summarized as follows:

\begin{itemize}
    \item We pioneer the use of both \textit{auxiliary} and the \textit{test-level covariates} for multiple hypothesis testing problems.
    \item We developed a novel MHT model that jointly learns \textit{test-level} and \textit{auxiliary covariates} through a neural network, which enables efficient optimization and gracefully handles high-dimensional hypothesis covariates. 

\end{itemize}

\section{Related work}
The traditional methods for controlling FDR, such as  Benjamini and Hochberg linear step-up procedure \cite{benjamini95}, and Storey's q value \cite{storey04} only use the $p$-values and impose the same threshold for all hypotheses. To increase the statistical power, many studies have been developed to take advantage of the \textit{test-level} information \cite{scott15,ignatiadis16,xia17,zhang19}. The general formulations considered in these papers assume that each hypothesis has an associated feature vector (or called \textit{test-level covariates}) related to the corresponding $p$-value. 

FDRreg \cite{scott15} 
adapts the two-groups model framework by taking into account the \textit{test-level covariates} information to model the mixing fraction in a regression setting. IHW \cite{ignatiadis16} groups the hypotheses into a pre-specified number of bins according to their associated feature space and applies a constant threshold for each bin to maximize the discoveries. One major limitation of IHW 
is that binning the data into groups can be tremendously difficult if the feature space is high-dimensional. NeuralFDR~\cite{xia17} addresses the limitation of IHW through the use of a neural network to parameterize the decision rule. This is a more general approach, and empirically it works well on a multi-dimensional feature space. AdaFDR \cite{zhang19} is an extension of NeuralFDR which models the discovery threshold by a mixture model using the expectation-maximization algorithm. The mixture model is a combination of a generalized linear model and Gaussian mixtures and displays improved power in comparison with IHW and NeuralFDR. However, AdaFDR only works with low-dimensional features, as its number of parameters 
grow linearly with respect to the covariate dimension. Thus, it is a substantial limitation for modern large-scale problems where a high-dimensional covariate setting is typical.

The recent work most relevant to ours is BB-FDR \cite{tansey18}. 
BB-FDR is the benchmark method for using a neural network to learn the true distributions of the test statistics from data in MHT. However, the existing model 
only deals with \textit{test-level covariates}, while our method enables the learning from both \textit{test-level covariates} and their associated \textit{auxiliary covariates}, and we formulated the model in a two-stage learning structure. Our method parametrizes the test-level covariates as a neural network and adjusts the feature hierarchy through a regression framework, which enables flexible handling of high-dimensional features as well as efficient end-to-end optimization.

\section{Preliminaries \label{sec:prel}}

Consider the situation with $n$ independent hypotheses whereby each hypothesis $i$ produces a test statistics $z_i$ corresponding to the test outcome. Now, each hypothesis also has $k$ test-level covariates $\mathbf{X}_{i} = (X_{i1},...,X_{ik})' \in \mathcal{R}^k$ and $q$ auxiliary  covariates $\mathbf{X}^a_{i} = (X^a_{i1},...,X^a_{iq})' \in \mathcal{R}^q$  characterized by a tuple $(z_i, \mathbf{X}_i,\mathbf{X}^a_i,  h_i)$,  where $h_i \in \{0,1\}$ indicates if the $i$th hypothesis is null ($h_i=0$) or alternative ($h_i=1$) which depends on both $\mathbf{X}_{i}$ and $\mathbf{X}^a_{i}$. The test statistics $z_i$ is calculated using data different from $\mathbf{X}_{i}$ and $\mathbf{X}^a_{i}$.  The standard assumption is that under the null ($h_i = 0$), the distribution of the test statistic $z_i$ is from the null distribution, denoted by $f_0(z)$; otherwise $z_i$ follows an unknown alternative distribution, denoted by $f_1(z)$. The alternative hypotheses for  $h_i = 1$ are the \emph{true signals} that we would like to discover.

The general goal of multiple hypotheses testing is to claim a maximum number of discoveries based on the observations $\{(z_i, \mathbf{X}_i, \mathbf{X}^a_i)\}_{i=1}^n$ while controlling the false positives. 
The most popular quantities that conceptualize the false positives are the family-wise error rate (FWER) \cite{dunn61} and the false discovery rate (FDR) \cite{benjamini95}.
We specifically consider FDR in this paper.
FDR is the expected proportion of false discoveries, and one closely related quantity, the false discovery proportion (FDP), is the actual proportion of false discoveries.
We note that FDP is the actual realization of FDR. 

\begin{figure*}[ht]
    \centering
    \includegraphics[width=0.65\textwidth]{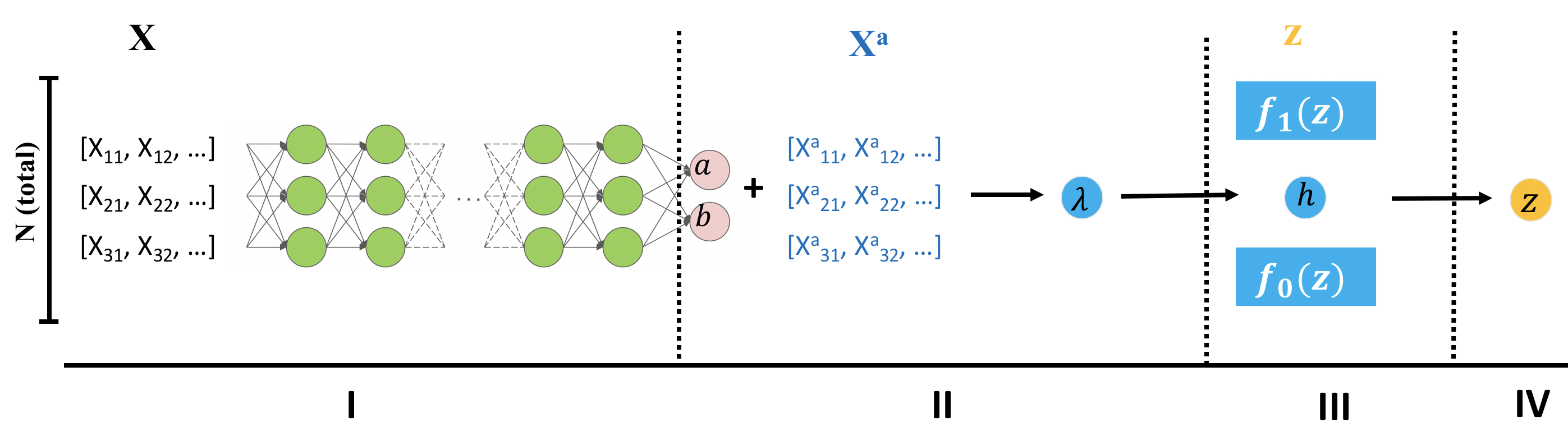}
    \caption{The graphical demonstration for NeurT-FDR. {I}: The deep neural network learning from the \textit{test-level covariates} ${\bf X}
    $; {II}: The bivariate linear regression adjustment on the beta parameters learned from the test level covariates, a, b, with the \textit{auxiliary covariates} ${\bf X}^a
    $. {III}: Mixing the bernoulli prior $h$ with the estimated alternative distribution of $f_1(z)$ and $f_0(z)$ from the input $z$; {IV}: The learned statistics $z$ from data.}
    \label{fig:model}
\end{figure*}

\subsection{False discovery rate control}
\label{subsec:background:fdr}
 For a given prediction $\hat{h}_i$, we say it is a true positive or a true discovery if $\hat{h}_i = 1 = h_i$ and a false positive or false discovery if $\hat{h}_i = 1 \neq h_i$. Let $\mathcal{D} = \{i : h_i = 1\}$ be the set of observations for which the treatment had an effect and $\hat{\mathcal{D}} = \{i : \hat{h}_i = 1\}$ be the set of predicted discoveries. We seek procedures that maximize the true positive rate (TPR) also known as \textit{power}, while controlling the false discovery rate -- the expected proportion of the predicted discoveries that are actually false positives.

\begin{definition} FDP and FDR

The false discovery proportion FDP and the false discovery rate FDR are defined as 
\begin{equation}
\label{eqn:fdp}
\text{FDR} \triangleq \coloneqq \mathbb{E}[\text{FDP}] \, , \quad \quad \text{FDP} \triangleq \frac{\#\{ i : i \in \hat{\mathcal{D}} \backslash \mathcal{D} \}}{\#\{ i : i \in \hat{\mathcal{D}} \}} \, .
\end{equation}
\end{definition}

In this paper, we aim to maximize ${\#\{ i : i \in \hat{\mathcal{D}} \}}$ while controlling $FDP\leq \alpha$ with high probability. 

\section{Method}\label{sec:method} 
\subsection{NeurT-FDR model description}
As shown in Figure \ref{fig:model},
NeurT-FDR extends the two-groups model \cite{efron08} and its hierarchical probabilistic extension \cite{tansey18} by learning a nonlinear mapping from the \textit{test-level covariates} (See Fig. \ref{fig:model}-I) and their associated \textit{auxiliary covariates} (See Fig. \ref{fig:model} II) jointly to model the test-specific mixing proportion (See Fig. \ref{fig:model}-III). More specifically, the model assumes a test-specific mixing proportion $\lambda_i$ which models the prior probability of the test statistics coming from the alternative (i.e. the probability of the test having an effect \textit{a priori}). Then, we place a Beta prior on each $\lambda_i$, as denoted in Eq. \ref{eqn:generative_model}. 

\begin{equation}
    \label{eqn:generative_model}
    \begin{aligned}
    z_i &\sim h_i f_1(z_i) + (1-h_i) f_0(z_i) \\ 
    h_i &\sim \text{Bernoulli}(\lambda_i) \\ 
    \lambda_i &\sim \text{Beta}(a_i, b_i),
    \end{aligned}
\end{equation}

Then, in order to borrow information from both $\mathbf{X}_i$  and $\mathbf{X}^a_i$ when inferencing on $\lambda_i$, ideally, one would estimate the parameters ($a_i$, $b_i$) of the Beta distribution from a neural network denoted by $G$ using the information of both $\mathbf{X}_i$  and $\mathbf{X}^a_i$ simultaneously. However,  the \textit{test-level covariates} are usually of complex high-dimensional features, while the auxiliary features are typically more structured low-dimensional, Thus, the information contained in the high-dimensional \textit{test-level covariates} may dominate the output of the deep neural network $G$~\cite{xu17}. Therefore, to better borrow information from the low-dimensional auxiliary features, we 
first learn a set of (pseudo) parameters, denoted by ($a'_i$, $b'_i$), of the Beta distribution with a deep neural network $G$ parameterized by $\theta_\phi$ from the high-dimensional \textit{test-level covariates} $\mathbf{X}$. Then, we further propose to adjust the learned pseudo parameters from the deep neural network through a linear regression on the auxiliary features $\mathbf{X}^a$ to determine the parameters for Beta distribution, denoted by ($a_i$, $b_i$) in Eq. (\ref{eqn:featureextraction1})  to (\ref{eqn:featureextraction2}).

\begin{align}
(a'_i, b'_i) &= G_{\theta_\phi}(\mathbf{X}_{i})  \label{eqn:featureextraction1} \\
\left[\begin{matrix}
\,
  log_e(a'_i)\;
\\[2\jot]
\hfill log_e(b'_i)
\,
\end{matrix} \right] &\sim \mathcal {N}_2 \left\{ \left[
  \begin{matrix}
    \mu_a + \mathbf{X}^{a}_i*\boldsymbol{\delta}_a \\ \mu_b + \mathbf{X}^{a}_i*\boldsymbol{\delta}_b
  \end{matrix}
  \right], \left[
  \begin{matrix}
    \sigma_{aa} & \sigma_{ab}  \\
    \sigma_{ab} & \sigma_{bb}\\
  \end{matrix}
  \right] \right\} \label{eqn:featureextraction2}
\end{align}

Notice that $\boldsymbol{\delta}_a$ and $\boldsymbol{\delta}_b$ are the coefficients of the bivariate linear regression. After fitting the bivariate linear regression on $\boldsymbol{X}^a$, we arrive at the fitted $\hat{a}'$ and $\hat{b}'$. Then we use the fitted mean value estimated from $\mu_a + X^a *\delta_a$, $\mu_b + X^b*\delta_b$ and the covariance matrix estimated from $cov(\hat{a}'_i - a'_i, \hat{b}'_i - b'_i)$ to generate the final adjusted $a_i, b_i$ from the bivariate normal distribution. 

\subsection{Learning Inference}
{We optimize $\theta_\phi$ by integrating out $h_i$ from Eq. (\ref{eqn:generative_model}) and maximizing the complete data log-likelihood as follows,
\begin{equation}
\label{eqn:bbfdr_data_likelihood}
\begin{aligned}
p_\theta(z_i) = \int_0^1 (\lambda_i f_1(z_i) + (1-\lambda_i) f_0(z_i)) \\ \times \text{Beta}(\lambda_i \vert \mathbf{X}_i, \mathbf{X}^a_i) d\lambda_i\,  .
\end{aligned}
\end{equation}

We opt for a beta prior because it is hierarchical and differently from other two-groups extensions, it uses a flatter hierarchy \cite{scott15,tansey18} improving training. First, optimization is easier and more stable because the output of the function is two soft-plus activations. Second, the additional hierarchy allows the model to assign different degrees of confidence to each test, changing the model from homoskedastic to heteroskedastic. 

We fit the model in Eq. \eqref{eqn:featureextraction1}, \eqref{eqn:featureextraction2} and \eqref{eqn:generative_model} with Stochastic Gradient Descent (SGD) on an $L_2$-regularized loss function,
\begin{equation}
\label{eqn:bbfdr_objective}
\begin{aligned}
& \underset{\theta \in \mathcal{R}^{|\theta|}}{\text{minimize}}
& & 
-\sum_{i} \log
p_\theta(z_i) + \lambda_i {G_{\theta_\phi}(\mathbf{X}_i)}_F^2 \, ,
\end{aligned}
\end{equation}

where ${\cdot}_F$ is the Frobenius norm. %
For computational purposes, we approximate the integral in Eq \eqref{eqn:bbfdr_data_likelihood} by a fine-grained numerical grid. Please check the Supplementary material for estimation details.

\begin{table*}[th]
\centering

\caption{Real data: \# of discoveries at FDR = 0.1. Best two performers per dataset are highlighted in bold.}
\label{tab:realdata}
\begin{tabular}{l|l|l|l|l}
          & Lapatinib  & Nutlin-3  & Airway    & Visual Tags\\
\hline
BH\citep{benjamini95}             & 117   & 151  & 4,079 & 312\\
SBH \citep{storey04}        &   131 (+11.9\%)   & 159 (+5.3\%)  & 4,079 & 312\\
AdaFDR \citep{zhang19}      & 137 (+9.7\%)  &   161 (+37.6\%)    & {\bf 6,050 (+48.3\%)} & -  \\
BB-FDR \citep{tansey18} & 181 (+54.7\%)  &  210 (+39.1\%)  & 5,791 (+41.9\%) & 385 (+23.4\%) \\
NeurT-FDRa (\textbf{ours})  &{\bf 187 (+59.8\%)}  & {\bf 215 (+42.3\%)}  &5,859 (+43.6\%) & {\bf 389 (+24.7\%)}\\
NeurT-FDRb (\textbf{ours})  & {\bf 212 (+81.2\%)}  & {\bf 260 (+72.2\%)}  &{\bf 8,820 (+116\%)} & {\bf 593 (+91.9\%)}\\
\end{tabular}
\end{table*}

\subsection{FDR control}
Once the optimized parameters $\hat{\theta}_\phi$ are chosen, we calculate the posterior probability of each test statistic coming from the alternative,
\begin{flalign}
\label{eqn:bbfdr_posterior}
\hat{w}_i &= p_{\hat{\theta}}(h_i = 1 | z_i) \\\nonumber
&= \int_0^1 \frac{\lambda_i f_1(z_i) \text{Beta}(\lambda_i | \mathbf{X}_i, \mathbf{X}^a_{i})}{\lambda_i f_1(z_i) + (1-\lambda_i) f_0(z_i)} d\lambda_i \, .
\end{flalign}

To maximize the total number of discoveries, first, we sort the posteriors in descending order by the likelihood of the test statistics being drawn from the alternative. We then reject the $m$ hypotheses, where $0 \leq m \leq n$ is the largest possible number such that the expected proportion of false discoveries is below the FDR threshold. Formally, this procedure solves the optimization problem,
\begin{equation}
\label{eqn:step_down_procedure_2}
\begin{aligned}
& \underset{m}{\text{maximize}}
& & 
m \\
& \text{subject to} & & \frac{\sum_{i=1}^m (1-\hat{w}_i)}{m} \leq \alpha \, ,
\end{aligned}
\end{equation}
for a given FDR threshold $\alpha$.

The neural network model $G$ uses the entire \textit{test-level} feature vector $X_{i\cdot}$ of every test to predict the prior parameters and then get adjusted by the entire \textit{auxiliary covariate} vector $X_{i\cdot}^{a}$ over $\lambda_i$. The observations $z_i$ are then used to calculate the posterior probabilities $\hat{w}_i$. The selection procedure in \eqref{eqn:step_down_procedure_2} uses these posteriors to reject a maximum number of null hypotheses while conserving the FDR.

\section{Case Studies\label{sec:emp}}
We evaluate our method \footnote{\href{https://github.com/lquvatexas/NeurT-FDR}{https://github.com/lquvatexas/NeurT-FDR}} using three real-world 
scenarios.  We consider BH \cite{benjamini95}, SBH \cite{storey04}, AdaFDR \cite{zhang19}, BB-FDR \cite{tansey18}, and two versions of our method, NeurT-FDRa, and NeurT-FDRb. For NeurT-FDRa, we only feed $\mathbf{X}$ into the $G_{\theta_\phi}$, while we stack $\mathbf{X}$ and $\mathbf{X}^a$ together and feed them into the $G_{\theta_\phi}$ for NeurT-FDRb.

{\bf Cancer drug screening data.}
 One goal of this analysis is to address the question of whether a given cell line responded to the drug treatment. Thus, this is a classical multiple testing problem that we need a hypothesis test for each cell line, where the null hypothesis is that the drug had no effect. We use the data preprocessed by \cite{tansey18} which contains genomic features and the z-score relative to mean control values for each cell line. We treat the genomic features as the \textit{test-level covariates} and extract the rank of the z-score as the \textit{auxiliary covariate}. For AdaFDR, we only use the \textit{auxiliary covariates} for the model input. Table \ref{tab:realdata} (columns Lapatinib and Nutlin-3) shows for both drugs NeurT-FDRa and NeurT-FDRb achieve the largest power compared to other methods and Figure \ref{fig:cancer_discoveries} shows that the \textit{test-level covariates} and \textit{auxiliary} features provide enough prior information that even some outcomes with a z-score above zero are still found to be significant in NeurT-FDRa. 
\begin{figure}[th]
    \centering
    \subfigure[\small BH on Lapatinib  \newline \text{  }(151 discoveries)]{\includegraphics[width=0.15\textwidth]{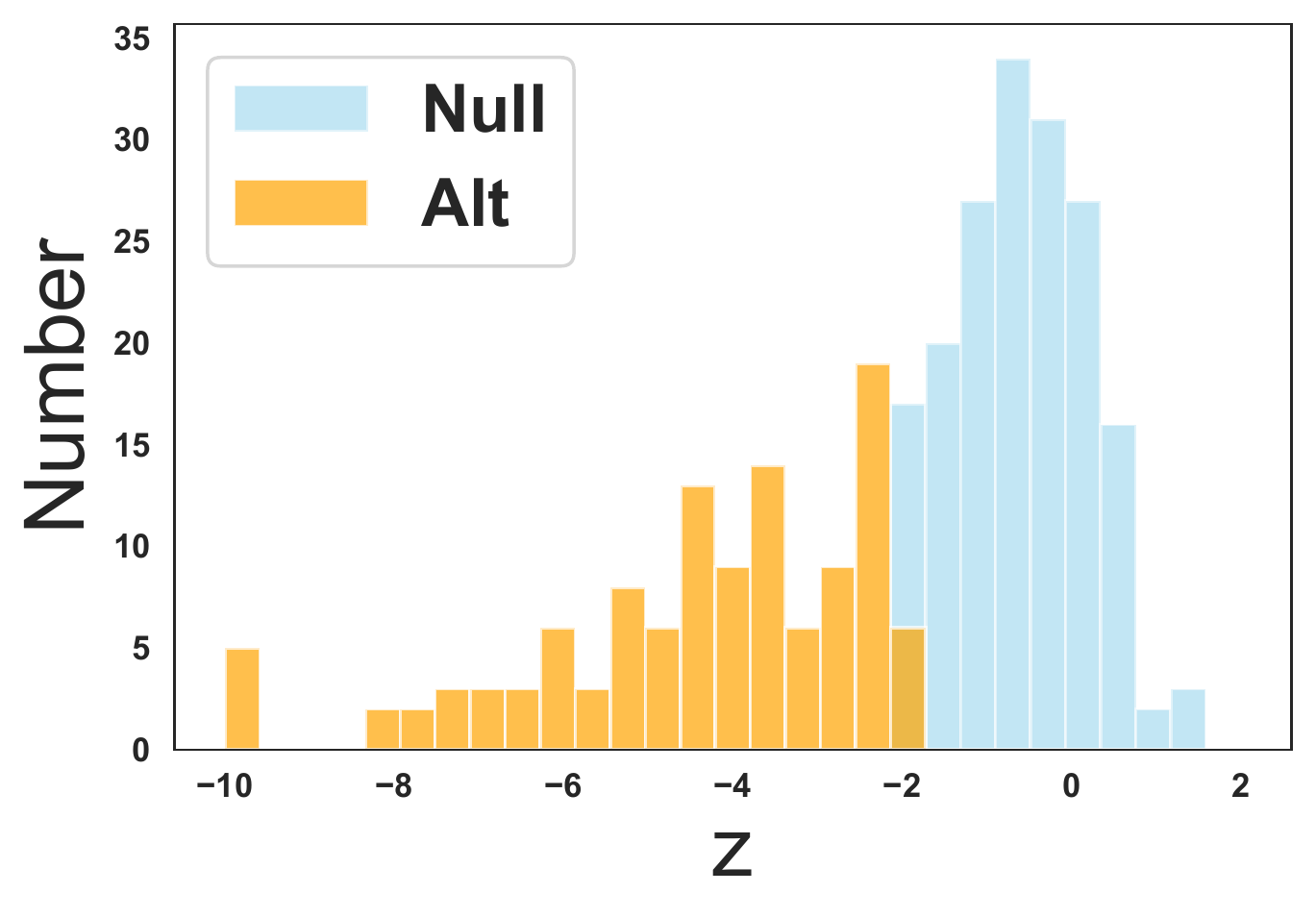}}
    \subfigure[\small BH on Nutlin-3 \newline  \text{  } (117 discoveries)]{\includegraphics[width=0.15\textwidth]{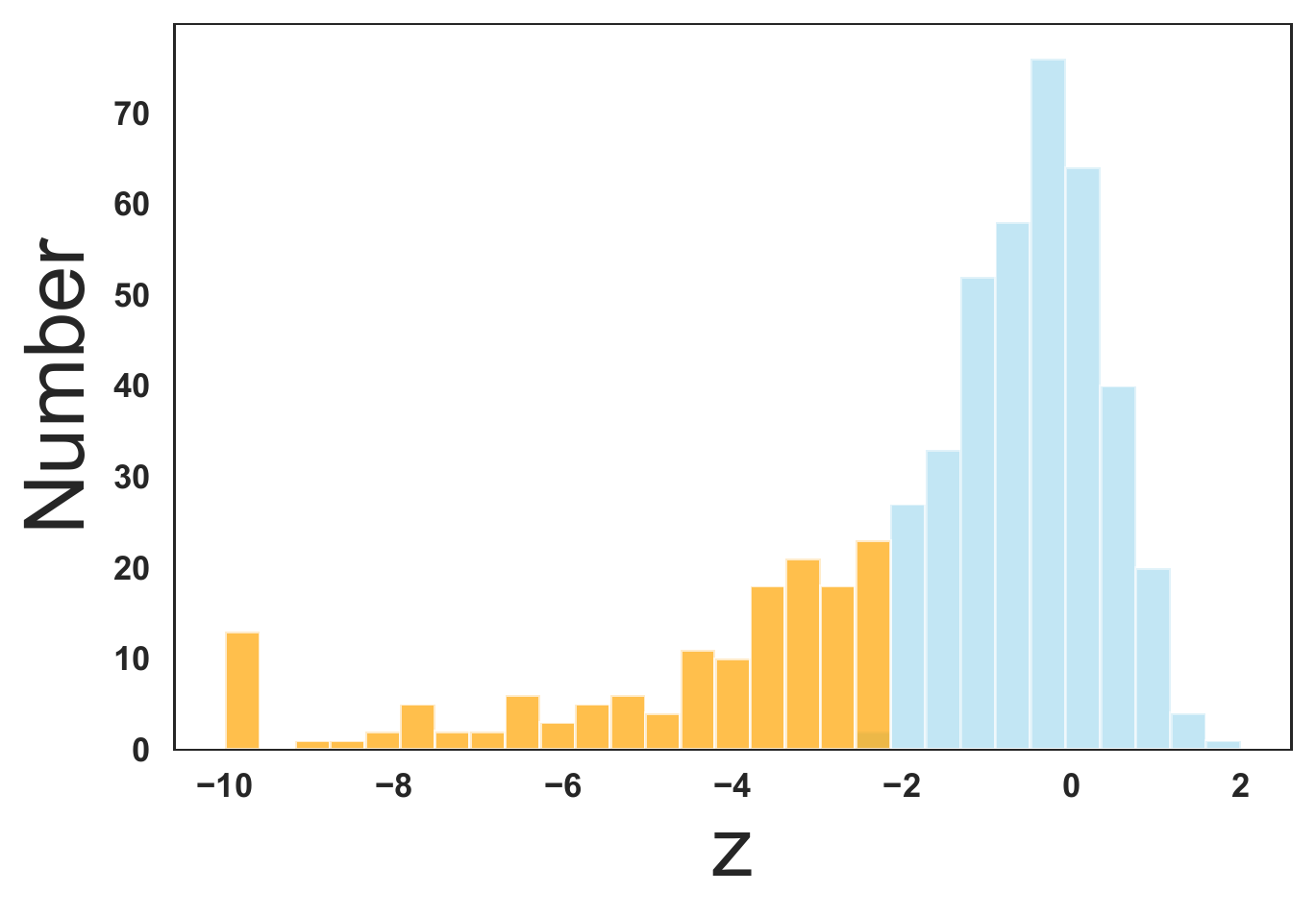}}
    
    \subfigure[\small NeurT-FDRa on Lapatinib \newline \text{  }(187 discoveries)]{\includegraphics[width=0.15\textwidth]{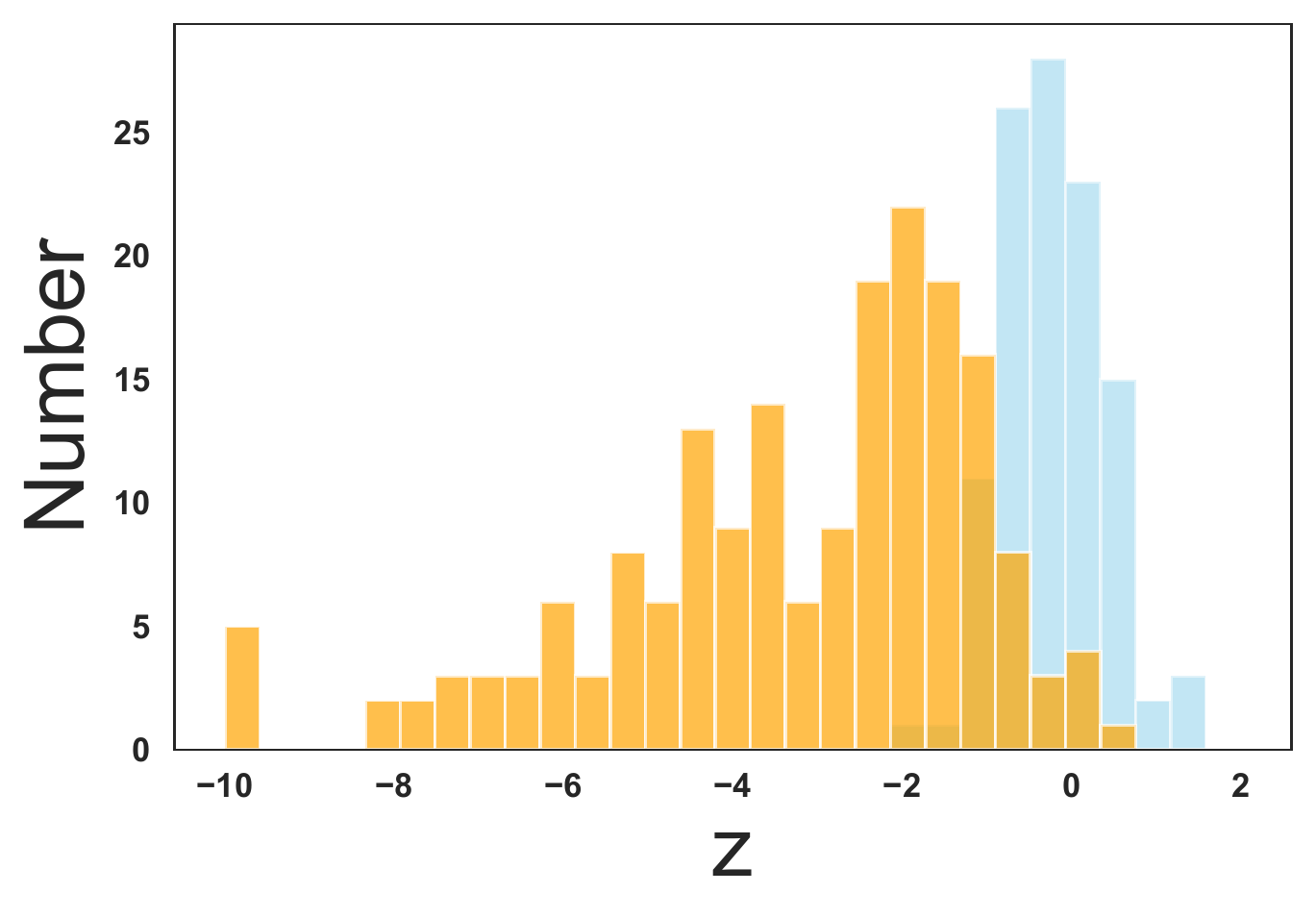}}
    \subfigure[\small NeurT-FDRa on Nutlin-3 \newline  \text{  } (215 discoveries)]{\includegraphics[width=0.15\textwidth]{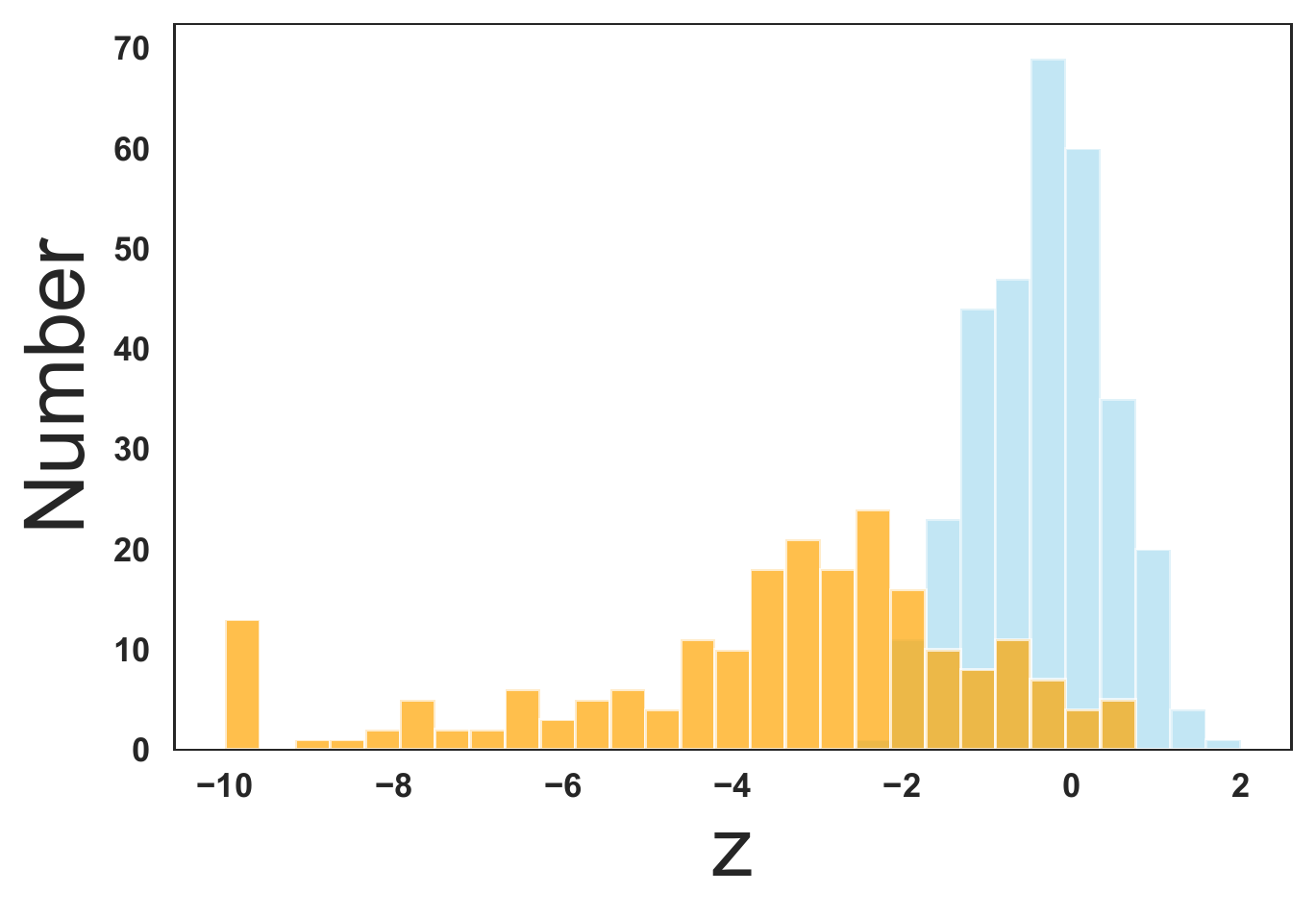}}
    \caption[Z-scores for the cancer drug case study]{\label{fig:cancer_discoveries} Discoveries found by NeurT-FDRa on the two drugs, compared to the discoveries found by a naive BH \cite{benjamini95} approach. Blue and orange represents the null and alternative discoveries respectively.}
\end{figure}

{\bf RNA-Seq data.}
 The original dataset contains a p-value and a log count for each gene (n=33,469), we consider the log count for each gene as the \textit{test-level covariate} and the rank for the p-value as the \textit{auxiliary covariate}. As the result shown in Table \ref{tab:realdata} (column Airway), where BB-FDR and NeurT-FDRa have a similar number of discoveries, AdaFDR performs slightly better and NeurT-FDRb provides 50\% more discoveries than all of them. All covariate-related methods make significantly more discoveries than the non-covariate-related methods. NeurT-FDRb achieves 116\% more discoveries compare to BH even when the dataset contains only one \textit{test-level covariate}.

{\bf Snap Visual Tags data.} 
 Each Snap has a visual tags vector coming from computer vision models with its corresponding content consumption metrics like how much time the particular user group spent on this Snap, number of shares, number of views, and others. So, we investigate which Snaps has the top engagements when they are compared to the normal behavior in one particular user cohort (i.e., age group and gender specification). We consider the visual tags as the \textit{test-level covariate} and the associated 16 content consumption metrics as the \textit{auxiliary covariates}. We used z-score as the ratio between Snap view time ratio and the number of view records to the mean values for each Snap. From our results in Table \ref{tab:realdata} (column visual tags), NeurT-FDRa and NeurT-FDRb provide significantly more discoveries than other methods, and here AdaFDR failed because it only can handle very low dimensions of covariates. AdaFDR worked in the cancer drug screening and RNA-seq data analysis when we used the rank of the test statistics as feature input. However, here we have 16 associated content consumption metrics which is a big advantage to our method since it is capable of handling both high-dimensional \textit{test} and \textit{auxiliary} level features' hypothesis test.

\section{Conclusion}
The neural network embedding architecture for the \textit{test-level covariates} and the linear regression model for learning the \textit{auxiliary covariates} enable NeurT-FDR to address modern high-dimensional problems. We believe NeurT-FDR will contribute to this field as a benchmark work for further investigation and have a wide application in 
neuroimaging, online advertising, 
and social media.
\bibliographystyle{ACM-Reference-Format}
\bibliography{cikm}


\begin{thebibliography}{11}


\ifx \showCODEN    \undefined \def \showCODEN     #1{\unskip}     \fi
\ifx \showDOI      \undefined \def \showDOI       #1{#1}\fi
\ifx \showISBNx    \undefined \def \showISBNx     #1{\unskip}     \fi
\ifx \showISBNxiii \undefined \def \showISBNxiii  #1{\unskip}     \fi
\ifx \showISSN     \undefined \def \showISSN      #1{\unskip}     \fi
\ifx \showLCCN     \undefined \def \showLCCN      #1{\unskip}     \fi
\ifx \shownote     \undefined \def \shownote      #1{#1}          \fi
\ifx \showarticletitle \undefined \def \showarticletitle #1{#1}   \fi
\ifx \showURL      \undefined \def \showURL       {\relax}        \fi
\providecommand\bibfield[2]{#2}
\providecommand\bibinfo[2]{#2}
\providecommand\natexlab[1]{#1}
\providecommand\showeprint[2][]{arXiv:#2}

\bibitem[Benjamini and Hochberg(1995)]%
        {benjamini95}
\bibfield{author}{\bibinfo{person}{Y. Benjamini} {and} \bibinfo{person}{Y.
  Hochberg}.} \bibinfo{year}{1995}\natexlab{}.
\newblock \showarticletitle{Controlling the false discovery rate: a practical
  and powerful approach to multiple testing}.
\newblock \bibinfo{journal}{\emph{Journal of The Royal Statistical Society.
  Series B (Methodological)}} \bibinfo{volume}{57}, \bibinfo{number}{1}
  (\bibinfo{year}{1995}), \bibinfo{pages}{289--300}.
\newblock


\bibitem[Dunn(1961)]%
        {dunn61}
\bibfield{author}{\bibinfo{person}{O.~J. Dunn}.}
  \bibinfo{year}{1961}\natexlab{}.
\newblock \showarticletitle{Multiple comparisons among means}.
\newblock \bibinfo{journal}{\emph{J. Amer. Statist. Assoc.}}
  \bibinfo{volume}{56}, \bibinfo{number}{293} (\bibinfo{year}{1961}),
  \bibinfo{pages}{52--64}.
\newblock


\bibitem[Efron(2004)]%
        {efron04}
\bibfield{author}{\bibinfo{person}{B. Efron}.} \bibinfo{year}{2004}\natexlab{}.
\newblock \showarticletitle{Large-scale simultaneous hypothesis testing: the
  choice of a null hypothesis}.
\newblock \bibinfo{journal}{\emph{J. Amer. Statist. Assoc.}}
  \bibinfo{volume}{99}, \bibinfo{number}{465} (\bibinfo{year}{2004}),
  \bibinfo{pages}{96 -- 104}.
\newblock


\bibitem[Efron(2008)]%
        {efron08}
\bibfield{author}{\bibinfo{person}{B. Efron}.} \bibinfo{year}{2008}\natexlab{}.
\newblock \showarticletitle{Microarrays, empirical bayes and the two-groups
  model}.
\newblock \bibinfo{journal}{\emph{Statist. Sci.}} \bibinfo{volume}{23},
  \bibinfo{number}{1} (\bibinfo{year}{2008}), \bibinfo{pages}{23 -- 28}.
\newblock


\bibitem[Ignatiadis et~al\mbox{.}(2016)]%
        {ignatiadis16}
\bibfield{author}{\bibinfo{person}{N. Ignatiadis}, \bibinfo{person}{J.~B.~Zaugg
  B.~Klaus}, {and} \bibinfo{person}{W. Huber}.}
  \bibinfo{year}{2016}\natexlab{}.
\newblock \showarticletitle{Data-driven hypothesis weighting increases
  detection power in genome-scale multiple testing}.
\newblock \bibinfo{journal}{\emph{Nature Methods}} \bibinfo{volume}{13},
  \bibinfo{number}{7} (\bibinfo{year}{2016}), \bibinfo{pages}{577--580}.
\newblock


\bibitem[Scott et~al\mbox{.}(2015)]%
        {scott15}
\bibfield{author}{\bibinfo{person}{J.~G. Scott}, \bibinfo{person}{R.~C. Kelly},
  \bibinfo{person}{M.~A. Smith}, \bibinfo{person}{P.~C. Zhou}, {and}
  \bibinfo{person}{R. E.}} \bibinfo{year}{2015}\natexlab{}.
\newblock \showarticletitle{False discovery rate regression: an application to
  neural synchrony detection in primary visual cortex}.
\newblock \bibinfo{journal}{\emph{Journal of American Statistical Association}}
  \bibinfo{volume}{110}, \bibinfo{number}{510} (\bibinfo{year}{2015}),
  \bibinfo{pages}{459 -- 471}.
\newblock


\bibitem[Storey et~al\mbox{.}(2004)]%
        {storey04}
\bibfield{author}{\bibinfo{person}{J.~D. Storey}, \bibinfo{person}{J.~E.
  Taylor}, {and} \bibinfo{person}{D. Siegmund}.}
  \bibinfo{year}{2004}\natexlab{}.
\newblock \showarticletitle{Strong control, conservative point estimation and
  simultaneous conservative consistency of false discovery rates: a unified
  approach.}
\newblock \bibinfo{journal}{\emph{Journal of The Royal Statistical Society.
  Series B (Methodological)}} \bibinfo{volume}{66}, \bibinfo{number}{1}
  (\bibinfo{year}{2004}), \bibinfo{pages}{187--205}.
\newblock


\bibitem[Tansey et~al\mbox{.}(2018)]%
        {tansey18}
\bibfield{author}{\bibinfo{person}{W. Tansey}, \bibinfo{person}{Y.~X. Wang},
  \bibinfo{person}{D.~M. Blei}, {and} \bibinfo{person}{R. Rabadan}.}
  \bibinfo{year}{2018}\natexlab{}.
\newblock \showarticletitle{Black Box FDR}. In
  \bibinfo{booktitle}{\emph{Proceedings of the 35th International Conference on
  Machine Learning (ICML 2018)}}. \bibinfo{pages}{4874--4883}.
\newblock


\bibitem[Xia et~al\mbox{.}(2017)]%
        {xia17}
\bibfield{author}{\bibinfo{person}{F. Xia}, \bibinfo{person}{M.~J. Zhang},
  \bibinfo{person}{J. Zou}, {and} \bibinfo{person}{D. Tse}.}
  \bibinfo{year}{2017}\natexlab{}.
\newblock \showarticletitle{NeuralFDR: Learning Discovery Thresholds from
  Hypothesis Features}. In \bibinfo{booktitle}{\emph{Proceedings of the 31th
  International Conference on Neural Information Processing Systems (NIPS
  2017)}}. \bibinfo{pages}{1540--1549}.
\newblock


\bibitem[Xu et~al\mbox{.}(2017)]%
        {xu17}
\bibfield{author}{\bibinfo{person}{J.L. Xu}, \bibinfo{person}{J.W. Han}, {and}
  \bibinfo{person}{F.P. Nie}.} \bibinfo{year}{2017}\natexlab{}.
\newblock \showarticletitle{Multi-view Feature Learning with Discriminative
  Regularization}. In \bibinfo{booktitle}{\emph{Proceedings of the 26th
  International Joint Conference on Artificial Intelligence (IJCAI-17)}}.
  \bibinfo{pages}{3161--3167}.
\newblock


\bibitem[Zhang et~al\mbox{.}(2019)]%
        {zhang19}
\bibfield{author}{\bibinfo{person}{M.~J. Zhang}, \bibinfo{person}{F. Xia},
  {and} \bibinfo{person}{J. Zou}.} \bibinfo{year}{2019}\natexlab{}.
\newblock \showarticletitle{Fast and covariate-adaptive method amplifies
  detection power in large-scale multiple hypothesis testing}.
\newblock
  \bibinfo{howpublished}{\url{https://doi.org/10.1038/s41467-019-11247-0}}.
\newblock \bibinfo{journal}{\emph{Nature Communications}}  \bibinfo{volume}{10}
  (\bibinfo{year}{2019}), \bibinfo{pages}{3433}.
\newblock
Issue 1.


\end{thebibliography}

\end{document}